# IPF for discrete chain factor graphs

Wim Wiegerinck and Tom Heskes
SNN, Department of Biophysics
Nijmegen University
P.O. Box 9101, 6500 HB, Nijmegen, the Netherlands
{wimw,tom}@mbfys.kun.nl

## Abstract

Iterative Proportional Fitting (IPF), combined with EM, is commonly used as an algorithm for likelihood maximization in undirected graphical models. In this paper, we present two iterative algorithms that generalize upon IPF. The first one is for likelihood maximization in discrete chain factor graphs, which we define as a wide class of discrete variable models including undirected graphical models and Bayesian networks, but also chain graphs and sigmoid belief networks. The second one is for conditional likelihood maximization in standard undirected models and Bayesian networks. In both algorithms, the iteration steps are expressed in closed form. Numerical simulations show that the algorithms are competitive with state of the art methods.

## 1 INTRODUCTION

Probabilistic graphical models are nowadays well established as modeling tool for multivariate domains with uncertainty [1, 2, 3]. The construction of a graphical model often consists of two steps. The first step is the determination of the network structure, which is usually done by hand on the basis of domain knowledge. In the second step, the network parameters are determined, usually by learning from a data set. This paper focuses on learning from data in a network with fixed structure, in a discrete variable setting.

A commonly used learning paradigm is maximum likelihood. In a Bayesian network with complete data, the maximum likelihood solution is easily found by counting conditional frequencies in the data. With incomplete data, e.g. due to missing data or to hidden variables, the EM algorithm [4] can straightforwardly be applied.

In undirected graphical models the Iterative Proportional Fitting (IPF) procedure is often used [5, 6, 7]. This is an iterative procedure with guaranteed convergence. It can be interpreted as cyclic coordinate-wise maximization of the likelihood. This method is also straightforwardly combined with EM to deal with incomplete data [8, 9].

Bayesian networks and undirected graphs are members of a larger family of probabilistic models that can be described as (locally) normalized products of local potentials. Other members that are of potential interest are chain graphs, that combine undirected models with directed models [7, 10], but also sigmoid belief networks [11]. These are parameterized Bayesian networks in which the conditional probabilities can be understood as locally normalized products of pairwise potentials between a child node and each of its parents. Parameterized networks are of particular importance for maximum likelihood applications, since they are more robust against over-fitting. In section 2, we define a wide model class that contains all the previously mentioned model classes, and combinations of them. The model class combines chain graphs and factor graphs [12], and we therefore call them 'chain factor graphs'.

In section 3 we derive an algorithm for maximum likelihood fitting of chain factor graphs. This algorithm generalizes upon EM in Bayesian networks and IPF/EM in undirected models. In particular, the generalized IPF can straightforwardly be applied to sigmoid belief networks, which to our knowledge has not been tried before in the literature. Simulations in a toy problem show favorable results compared to maximization with conjugate gradient maximization (section 5).

Graphical models, in particular Bayesian networks, are often found in diagnostic applications. In such applications data sets are often conditional [13, 14]. For instance, a hospital data set on a certain disease is not unlikely to be conditioned on patients complaints.

For conditional data, maximum conditional likelihood is an appropriate learning paradigm. In section 4, we derive an IPF-like algorithm fitting of maximum conditional likelihood for a restricted class of chain factor graph models. This model class includes Bayesian networks and undirected graphs. In section 5, this algorithm is applied to a toy problem with real medical conditional data. It shows favorable results compared to the recently developed 'reverse Jensen' algorithm for conditional likelihood maximization [15].

## 2  CHAIN FACTOR GRAPHS

In this section, we describe a general class of probabilistic models that are parameterized as locally normalized products of cluster potentials. To represent these models, we combine chain graphs (e.g. [10, 7]) with factor graphs [12], hence the name chain factor graphs.

In this article we restrict ourselves to probabilistic models $P(x)$ on a finite set of categorical variables, i.e. $x = (x_1, \ldots, x_n)$ with $x_i \in \{1, \ldots, n_i\}$. We often use the sets as lower indices to denote sub-vectors of $x$ (see e.g. [6]). So if $\alpha = \{\alpha_1, \ldots, \alpha_n\}$, our notation is $x_\alpha \equiv \{x_{\alpha_1}, \ldots, x_{\alpha_n}\}$.

In our definition, a chain factor graphical model satisfies a component factorization

$$P(x) = \prod_A P(x_{C(A)}|x_{\pi(A)}) \quad (1)$$

in which $A$ are possibly overlapping sets that each partition into a non-empty subset $C(A)$ -called the chain component- and its (possibly empty) complement $\pi(A) = A\setminus C(A)$ - called the parent of $C(A)$. The chain components $C(A)$ should form a mutually disjoint partition of the set of all variables $\{1\ldots n\}$. The sets $A$ should allow an ordering such that parent-sets $\pi(A)$ are contained in preceding chain components,

$$\pi(A) \subset \bigcup_{B<A} C(B) \quad (2)$$

Each of the conditional distributions $P(x_{C(A)}|x_{\pi(A)})$ in a chain factor graph is a conditional factor graph defined by non-negative subcluster potentials $\psi_\alpha$, with possibly overlapping subclusters $\alpha$, i.e.

$$P(x_{C(A)}|x_{\pi(A)}) = \frac{1}{Z_A(x_{\pi(A)})} \prod_{\alpha \to A} \psi_\alpha(x_\alpha) \quad (3)$$

which is normalized by $Z_A(x_{\pi(A)})$,

$$Z_A(x_{\pi(A)}) = \sum_{x_{C(A)}} \prod_{\alpha \to A} \psi_\alpha(x_\alpha) \quad \forall x_{\pi(A)}. \quad (4)$$

The notation $\alpha \to A$ indicates that $\alpha$ is a 'member' of $A$. The member clusters should cover $A$, i.e., $\bigcup_{\alpha \to A} \alpha = A$. Unlike the standard definition of chain graphs as in e.g. [10, 7], there need not be a cluster $\alpha \to A$ that covers $\pi(A)$. An example of such a model is the sigmoid belief network (see below).

**Examples.** An undirected model is a chain factor graph with only one set $A = \{1, \ldots, n\}$ that consists of all the nodes. Therefore $C(A) = A$ and $\pi(A) = \emptyset$. The set of subclusters $\alpha$ may be any set of clusters covering $\{1, \ldots, n\}$. If the subclusters $\alpha$ are cliques in an undirected graph, the model is a Markov network. If the subclusters $\alpha$ are the pairs of nodes that are linked in an undirected graph, the model corresponds to a Boltzmann machine.

A directed graphical model is a chain factor graph in which each set $A$ coincides with a node $i$ and its parents $pa(i)$ in the directed graph: $C(A) = i$, $\pi(A) = pa(i)$. In a conventional Bayesian network, $A$ has only one member which coincides with $A$.

A sigmoid belief network with binary $(-1, 1)$ units is a directed graphical model in which the conditional probability of a node $i$ given its parents $pa(i)$ is usually parameterized by

$$P(x_i = 1|x_{pa(i)}) = \sigma(2(h_i + \sum_{j \in pa(i)} w_{ij}x_j)) \quad (5)$$

where $\sigma(x) = \exp(x)/(1 + \exp(x))$. The parameters $w_{ij}$ are called the weights and $h_i$ the biases. In this model, $A = \{i, pa(i)\}$, $C(A) = i$, $\pi(A) = pa(i)$. The clusters $\alpha$ are pairs $\{i, j\}$, with $j \in pa(i)$. Weights and biases can be transformed into pair-wise potentials via $\psi_{ij}(x_i, x_j) \equiv \exp(x_i[w_{ij}x_j - h_i/k_i])$, where the $k_i$ must be such that they add up to the number of parents. Conversely, any set of pair-wise potentials $\psi_{ij}$ between node $i$ and its parents can be transformed into unique weights and biases.

## 3  MAXIMUM LIKELIHOOD

In this section, we derive an algorithm for maximum likelihood estimation for chain factor graphs with a data set of $N$ independent observations $D = \{x^\mu_{v(\mu)}\}$. In the data, the label $\mu$ indexes the observations. The subindex $v(\mu)$ labels the domain of the observations. In general, maximum likelihood estimation tries to find model parameters $\psi$ that maximize the log-likelihood function

$$L(\psi) = \frac{1}{N} \sum_\mu \log P(x^\mu_{v(\mu)}|\psi). \quad (6)$$

The most obvious way to proceed is trying to maximize $L$ directly by applying standard algorithms for nonlinear optimization. A widely applied alternative method is to construct an auxiliary function $F$ which bounds $L$ from below. If we then take a step in parameter space that maximizes $F$, instead of $L$, $L$ is incremented as well. This standard technique can be viewed as the basis for e.g. the EM algorithm [4] and iterative scaling [16].

Before we construct an auxiliary function for the log-likelihood in chain factor graphs, we first review (coordinate-wise) auxiliary functions in a more general setting. We assume that the parameters $\psi$ can be written as a set of potentials $\psi_\alpha$, and we write $\psi_{-\alpha}$ for the potentials excluding $\psi_\alpha$, i.e., $\psi = (\psi_\alpha, \psi_{-\alpha})$. We call $F_\alpha$ a coordinate-wise auxiliary function for $L$ if for any setting of $\psi_\alpha$, $\widetilde{\psi}_\alpha$, and $\widetilde{\psi}_{-\alpha}$ it has the following two properties:

$$\text{I} \quad F_\alpha(\psi_\alpha; \widetilde{\psi}) - F_\alpha(\widetilde{\psi}_\alpha; \widetilde{\psi}) \leq L(\psi_\alpha, \widetilde{\psi}_{-\alpha}) - L(\widetilde{\psi}_\alpha, \widetilde{\psi}_{-\alpha}) \quad (7)$$

$$\text{II} \quad \left.\frac{\partial F_\alpha(\psi_\alpha; \widetilde{\psi})}{\partial \psi_\alpha}\right|_{\psi_\alpha = \widetilde{\psi}_\alpha} = \left.\frac{\partial L(\psi)}{\partial \psi_\alpha}\right|_{\psi = \widetilde{\psi}} \quad (8)$$

If we have coordinate wise auxiliary functions for all $\alpha$, we can monotonically increase $L$ by Algorithm 1.

---

**Algorithm 1** Maximum likelihood with auxiliary function

1: initialize($\widetilde{\psi}$)
2: **repeat**
3:   **for_all** $\alpha$ **do**
4:     $\widetilde{\psi}_\alpha \leftarrow \arg\max_{\psi_\alpha} F_\alpha(\psi_\alpha; \widetilde{\psi})$
5:   **end for**
6: **until** convergence criterion is met
7: **return** $\widetilde{\psi}$

---

Property I guarantees that each step of Algorithm 1 increases $L$. Property II guarantees that the procedure is only stationary in points where the gradient of $L$ is zero. Although the method is arguably elegant, the method can only be expected to provide better performance than standard gradient based optimization techniques if both following conditions hold:

1. The computation of arg max in line 4 in Algorithm 1 is much more efficient than line search in $L$.

2. The bound in property I is tight. Otherwise the procedure reduces to a coordinate gradient ascent-method.

We now describe how to derive coordinate-wise EM as a special case of Algorithm 1. With incomplete data, the log-likelihood (6) is

$$L(\psi) = \frac{1}{N} \sum_\mu \log \sum_{x_{h(\mu)}} P(x_{h(\mu)}, x_{v(\mu)}^\mu | \psi) \quad (9)$$

where $h(\mu)$ is the complement of $v(\mu)$. Using Jensen's inequality, $f(E(x)) \geq E(f(x))$ for concave $f$ such as $\log x$, we can derive coordinate-wise EM auxiliary functions,

$$F_\alpha^{EM}(\psi_\alpha; \widetilde{\psi}) = \frac{1}{N} \sum_\mu \sum_{x_{h(\mu)}} \left[ P(x_{h(\mu)} | x_{v(\mu)}^\mu, \widetilde{\psi}) \right.$$
$$\left. \times \log P(x_{h(\mu)}, x_{v(\mu)}^\mu | \psi_\alpha, \widetilde{\psi}_{-\alpha}) \right]$$
$$= \sum_x \widetilde{P}^D(x) \log P(x | \psi_\alpha, \widetilde{\psi}_{-\alpha}), \quad (10)$$

where $\widetilde{P}^D$ is the probability of $x$ given the data $D$ and the old parameters $\widetilde{\psi}$,

$$\widetilde{P}^D(x) \equiv \frac{1}{N} \sum_\mu P(x_{h(\mu)} | x_{v(\mu)}^\mu, \widetilde{\psi}) I_{x_{v(\mu)}^\mu}(x_{v(\mu)}) \quad (11)$$

and $I_{x_a'}(x)$ is the indicator function, which is one if $x_a = x_a'$ and zero otherwise. The EM auxiliary function has turned the incomplete data log-likelihood into a complete data log-likelihood, where the complete data is distributed according to $\widetilde{P}^D$.

For chain factor graphs, the coordinate wise EM-auxiliary function boils down to

$$F_\alpha^{\text{EM}}(\psi_\alpha; \widetilde{\psi}) = \sum_{x_\alpha} \widetilde{P}^D(x_\alpha) \log \psi_\alpha(x_\alpha)$$
$$- \sum_{x_{\pi(A)}} \widetilde{P}^D(x_{\pi(A)}) \log Z_A(x_{\pi(A)}) \quad (12)$$

where $Z_A$ is the normalization factor for the chain with member $\alpha$, see (4). In the context here, it is important to note that $Z_A$ is a function of $\psi_\alpha$,

$$Z_A(x_{\pi(A)}) = \sum_{x_{C(A)}} \psi_\alpha(x_\alpha) \widetilde{\Psi}_{A \backslash \alpha}(x_A) \quad (13)$$

with $\widetilde{\Psi}_{A \backslash \alpha}(x_A) \equiv \prod_{\beta \to A; \beta \neq \alpha} \widetilde{\psi}_\beta(x_\beta)$. In particular due to the $\log Z_A$ term, the EM-auxiliary function (12) is still too inconvenient to use in Algorithm 1. From the concavity of log, we have the additional bound,

$$\log Z_A \leq \frac{Z_A}{\widetilde{Z}_A} + \log \widetilde{Z}_A - 1 \quad (14)$$

where $\widetilde{Z}_A$ is the normalization factor for the parameters $\psi_\alpha = \widetilde{\psi}_\alpha$. This bound of $\log Z_A$ is linear in $\psi_\alpha$.

Using this bound, we can derive the much more convenient auxiliary functions

$$F_\alpha(\psi_\alpha; \widetilde{\psi}) = \sum_{x_\alpha} \widetilde{P}^D(x_\alpha) \log \psi_\alpha(x_\alpha) - \sum_{x_\alpha} \widetilde{g}_\alpha(x_\alpha) \psi_\alpha(x_\alpha) \quad (15)$$

where

$$\widetilde{g}_\alpha(x_\alpha) = \sum_{x'_A} \widetilde{P}^D(x'_{\pi(A)}) \frac{\widetilde{\Psi}_{A\setminus\alpha}(x'_A) I_{x_\alpha}(x'_A)}{\widetilde{Z}_A(x'_{\pi(A)})}. \quad (16)$$

The argmax, needed in Algorithm (1) can now easily be solved in closed form,

$$\operatorname*{argmax}_{\psi_\alpha} F_\alpha(\psi_\alpha; \widetilde{\psi}) = \frac{\widetilde{P}^D(x_\alpha)}{\widetilde{g}_\alpha(x_\alpha)}. \quad (17)$$

We will show below that the resulting algorithm generalizes upon iterative proportional fitting (IPF). In the remainder of the paper, we will therefore call this algorithm, and similar ones that are derived in the next section, generalized IPF or simply IPF.

**Examples.** (I) In Bayesian networks, a set $A$ has only one member $A = \alpha$. This has two consequences: $\widetilde{\Psi}_{A\setminus\alpha}(x'_A) \equiv 1$, and $\pi(A) \subset \alpha$. Therefore $\widetilde{g}_\alpha(x_\alpha) = [\widetilde{P}^D(x_{\pi(A)})][\widetilde{Z}_A(x_{\pi(A)})]^{-1} \equiv \widetilde{z}(x_{\pi(A)})$. If we substitute the solution $\psi_\alpha^*$ of (17) into the updated conditional probability $P^*(x_{C(A)}|x_{\pi(A)})$ of (3), we find that $\widetilde{z}$ drops out. With $C(A) = i$ and $\pi(A) = \text{pa}(i)$ we recover the familiar EM update rule

$$P^*(x_i|x_{\text{pa}(i)}) = \widetilde{P}^D(x_i|x_{\text{pa}(i)}). \quad (18)$$

(II) In an undirected model, there is only one set $A$. Therefore $\pi(A) = \emptyset$, and $\widetilde{P}^D(x'_{\pi(A)})[\widetilde{Z}_A(x'_{\pi(A)})]^{-1} \equiv z$ is an irrelevant constant. The solution $\psi_\alpha^*$ of (17) is

$$\psi_\alpha^*(x_\alpha) = \frac{\widetilde{P}^D(x_\alpha)}{z \sum_{x'} \widetilde{\Psi}_{-\alpha}(x') I_{x_\alpha}(x')} \quad (19)$$

with $\widetilde{\Psi}_{-\alpha} = \prod_{\beta \neq \alpha} \widetilde{\psi}_\beta$. Let us consider how this can be expressed as an update rule for $P$. We start with

$$\widetilde{P}(x) \propto \widetilde{\psi}_\alpha(x_\alpha) \widetilde{\Psi}_{-\alpha}(x). \quad (20)$$

The marginal in cluster $\alpha$ according to $\widetilde{P}$ is

$$\widetilde{P}(x_\alpha) \propto \widetilde{\psi}_\alpha(x_\alpha) \sum_{x'} \widetilde{\Psi}_{-\alpha}(x') I_{x_\alpha}(x'). \quad (21)$$

So the conditional distribution $\widetilde{P}(x_{-\alpha}|x_\alpha)$ is

$$\widetilde{P}(x_{-\alpha}|x_\alpha) \propto \frac{\widetilde{\Psi}_{-\alpha}(x)}{\sum_{x'} \widetilde{\Psi}_{-\alpha}(x') I_{x_\alpha}(x')}, \quad (22)$$

which is independent of the potential $\widetilde{\psi}_\alpha$. So updating $\widetilde{\psi}_\alpha$ leaves $\widetilde{P}(x_{-\alpha}|x_\alpha)$ invariant. On the other hand, if we consider the updated probability $P^*(x) \propto \psi_\alpha^* \widetilde{\Psi}_{-\alpha}$, we find that it fits the margin of the EM-estimated data on cluster $\alpha$:

$$P^*(x_\alpha) = \widetilde{P}^D(x_\alpha). \quad (23)$$

So

$$P^*(x) = \widetilde{P}(x_{-\alpha}|x_\alpha) \widetilde{P}^D(x_\alpha), \quad (24)$$

which is exactly the update rule of IPF (see e.g. [6]).

In models where the parent set $\pi(A)$ of each $A$ is contained in at least one $\alpha \to A$ (as in e.g. chain graphs), elaboration of the IPF solution (17) shows that it can be reduced to recursively applying IPF to the undirected models $Z_A^{-1} \prod_{\alpha \to A} \psi_\alpha(x_\alpha)$ on $A$. In these recursively obtained solutions, however, one should take the completed distributions $\widetilde{P}^D$ based on the full model. This recursive application of IPF is described in e.g. [7] to maximize the likelihood in chain graphs.

## 4 MAXIMUM CONDITIONAL LIKELIHOOD

In this is section we derive IPF-like algorithms for maximum conditional likelihood fitting of chain factor graphs to a conditional data set $D^c = \left\{x^\mu_{v(\mu)}, x^\mu_{c(\mu)}\right\}_\mu$. In this data set the observation $x^\mu_{v(\mu)}$ has been made conditioned on the variables in $c(\mu)$ which are clamped in the states $x^\mu_{c(\mu)}$. The sub-indices $v(\mu)$ and $c(\mu)$ indicate the domains of the observed and the clamped variables respectively. In general, maximum conditional likelihood tries to find model parameters $\psi$ such that the conditional log-likelihood

$$L^c = \frac{1}{N} \sum_\mu \log P(x^\mu_{v(\mu)}|x^\mu_{c(\mu)}, \psi)$$
$$= \frac{1}{N} \sum_\mu \log P(x^\mu_{v(\mu)}, x^\mu_{c(\mu)}|\psi) - \frac{1}{N} \sum_\mu \log P(x^\mu_{c(\mu)}|\psi) \quad (25)$$

is maximized. In analogy with the previous section, our approach is to derive a set of coordinate wise auxiliary functions $F_\alpha^c$ for $L^c$. If we then replace $F_\alpha$ by $F_\alpha^c$ in Algorithm 1, we can apply the algorithm directly for conditional likelihood maximization. Unfortunately, the tools that we used in the previous section are, in general, not sufficiently powerful to bound $-\sum_\mu \log P(x^\mu_{c(\mu)})$ in a convenient way. However, there are two subclasses of problems where we can proceed.

## 4.1 CLAMPED PARENTS

The first case, which is the easiest one, is where the argument $x_{\pi(A)}$ of the normalization factors $Z_A$ (see (4)) are clamped, i.e., $\pi(A) \subset c(\mu)$. (This is always the case for an undirected factor graph). In that case, all the $Z_A$ drop out of (25). The remaining conditional log-likelihood for chain factor graphs is

$$L^c = \frac{1}{N}\sum_\mu \log \sum_x I_{x^\mu_{v,c}}(x)\Psi_{-\alpha}(x)\psi_\alpha(x_\alpha)$$
$$-\frac{1}{N}\sum_\mu \log \sum_x I_{x^\mu_c}(x)\Psi_{-\alpha}(x)\psi_\alpha(x_\alpha) \quad (26)$$

where we have defined $\Psi_{-\alpha} = \prod_{\beta \neq \alpha} \psi_\beta$. The first term in (26) is treated using Jensen's bound as in standard EM. For the second term, which is due to the normalization of $P(x^\mu_v|x^\mu_c)$, we use the linear bound of log as in (14). This is similar to the way we dealt with the normalization factor $Z_A$ in the previous section. Thus we obtain auxiliary functions for the conditional log-likelihood $L^c$ that are very similar to the ones for the unconditional log-likelihood (cf. (15))

$$F^c_\alpha(\psi_\alpha; \widetilde{\psi}) = \sum_{x_\alpha} \widetilde{P}^D(x_\alpha) \log \psi_\alpha(x_\alpha)$$
$$- \sum_{x_\alpha} \widetilde{g}^c_\alpha(x_\alpha)\psi_\alpha(x_\alpha). \quad (27)$$

In (27), $\widetilde{P}^D$ is the probability over $x$ given the data $D^c$ as if it would be unconditional data, and given the old parameters $\widetilde{\psi}$. Furthermore,

$$\widetilde{g}^c_\alpha(x_\alpha) = \sum_\mu \frac{\sum_{x'} I_{x^\mu_c}(x')\widetilde{\Psi}_{-\alpha}(x')I_{x_\alpha}(x')}{\sum_{x'} I_{x^\mu_c}(x')\widetilde{\Psi}_{-\alpha}(x')\widetilde{\psi}_\alpha(x'_\alpha)}. \quad (28)$$

The solution of argmax needed in line 4 of Algorithm 1 for conditional data now is

$$\arg\max_{\psi_\alpha} F^c_\alpha(\psi_\alpha; \widetilde{\psi}) = \frac{\widetilde{P}^D(x_\alpha)}{\widetilde{g}^c_\alpha(x_\alpha)}. \quad (29)$$

## 4.2 JOINT PARENTS

The second case is more complex. Here the clamped distribution is unrestricted, but now the chain factor graphs are restricted. We only consider chain factor graphs in which each cluster $\alpha$ contains all the parents of the chain component: $\alpha \to A$ implies $\pi(A) \subset \alpha$. In a way, the parents are always together, hence the name "joint parents". Examples are Bayesian networks, and more general, directed cluster graphs in which the clusters $A$ are not subdivided into more subclusters. Obviously, sigmoid belief networks do not fall in this subclass. Undirected models have no parents and thus (trivially) "joint" parents, but for those the 'clamped parents' procedure is more appropriate.

With joint parents, we can get rid of the normalization factor by absorbing it in any of the related potentials, i.e. if the normalization is $Z_A(x_{\pi(A)})$, we can obtain an equivalent model with normalization $Z'_A(x_{\pi(A)}) = 1$ by multiplying any of the potentials $\psi_\alpha(x_\alpha)$ by $[Z_A(x_{\pi(A)})]^{-1}$. As a consequence, we may constrain the potentials in a joint parent chain factor graph by $Z_A(x_{\pi(A)}) = 1$, without the probabilistic models being constrained.

By constraining the normalization factors to 1, the conditional log-likelihood for joint parent chain factor graphs reduces to (26). However, now (26) is to be maximized under the linear constraints

$$\sum_{x_{C(A)}} \psi_\alpha(x_\alpha)\widetilde{\Psi}_{A\setminus\alpha}(x_A) = 1. \quad (30)$$

We still can use the method of auxiliary functions, using (27). However, we should adapt Algorithm 1 such that the constraints are initially satisfied (line 1) and remain satisfied (line 4). In particular, the optimization (line 4 in Algorithm 1) needs some attention. Since $F^c_\alpha$ is concave in $\psi_\alpha$, and since the constraints (30) are linear in $\psi_\alpha$, we can apply the theory of constraint optimization with Lagrange multipliers [17]. We introduce Lagrange multipliers $\lambda(x_{\pi(A)})$ and define the Lagrangian

$$\widetilde{\mathcal{F}}_\alpha(\psi_\alpha, \lambda; \widetilde{\psi}) = F^c_\alpha(\psi_\alpha; \widetilde{\psi})$$
$$- \sum_{x_{\pi(A)}} \lambda(x_{\pi(A)})[\sum_{x_{C(A)}} \psi_\alpha(x_\alpha)\widetilde{\Psi}_{A\setminus\alpha}(x_A) - 1]$$
$$= F^c_\alpha(\psi_\alpha; \widetilde{\psi}) - \sum_{x_\alpha} \lambda(x_{\pi(A)})\widetilde{h}_\alpha(x_\alpha)\psi_\alpha(x_\alpha)$$
$$+ \sum_{x_{\pi(A)}} \lambda(x_{\pi(A)}) \quad (31)$$

with

$$\widetilde{h}_\alpha(x_\alpha) \equiv \sum_{x'_A} \widetilde{\Psi}_{A\setminus\alpha}(x'_A)I_{x_\alpha}(x'_A) \quad (32)$$

which should be maximized with respect to $\psi_\alpha$ and minimized with respect to $\lambda$. Maximizing with respect $\psi_\alpha$ for fixed $\lambda$ yields

$$\psi^\lambda_\alpha(x_\alpha) = \frac{\widetilde{P}^D(x_\alpha)}{\widetilde{g}^c_\alpha(x_\alpha) + \lambda(x_{\pi(A)})\widetilde{h}_\alpha(x_\alpha)}. \quad (33)$$

The optimal $\lambda$ can be found by minimizing the dual function $\widetilde{\mathcal{F}}_\alpha(\psi^\lambda_\alpha, \lambda; \widetilde{\psi})$ or simply by searching $\lambda$ such that it matches the constraints, i.e.,

$$\sum_{x_{C(A)}} \psi_\alpha^\lambda(x_\alpha)\widetilde{\Psi}_{A\backslash\alpha}(x_A) \equiv \mathcal{Z}_\alpha(\lambda; x_{\pi(A)}) = 1 \quad (34)$$

for all $x_{\pi(A)}$. Since $\pi(A) \subset \alpha$, the constraints decouple in the components $\lambda(x_{\pi(A)})$, i.e. $\mathcal{Z}_\alpha(\lambda; x_{\pi(A)}) = \mathcal{Z}_\alpha(\lambda(x_{\pi(A)}); x_{\pi(A)})$. For each $x_{\pi(A)}$, this function is monotonically decreasing in $\lambda(x_{\pi(A)})$. Therefore, a solution $\lambda(x_{\pi(A)})$ that satisfies (34) for $x_{\pi(A)}$ within some precision $\epsilon$, can be found efficiently with bisection [18]. Note that $\widetilde{P}^D(x_\alpha)$, $\widetilde{g}_\alpha^c(x_\alpha)$, and $\widetilde{h}_\alpha(x_\alpha)$ are independent of $\lambda$ and therefore do not have to be recomputed during this search. Finally, we remark that we can bracket $\lambda(x_{\pi(A)})$ initially by

$$\lambda_{\min}^0(x_{\pi(A)}) \equiv \max_{x_{C(\alpha)}} \widetilde{P}^D(x_\alpha)(\widetilde{h}_\alpha(x_\alpha))^2 - \frac{\widetilde{g}_\alpha^c(x_\alpha)}{\widetilde{h}_\alpha(x_\alpha)} \quad (35)$$

$$\lambda_{\max}^0(x_{\pi(A)}) \equiv \sum_{x_{C(\alpha)}} \widetilde{P}^D(x_\alpha) \quad (36)$$

since substitution shows that $\mathcal{Z}_\alpha(\lambda_{\min}^0; x_{\pi(A)}) \geq 1$ and $\mathcal{Z}_\alpha(\lambda_{\max}^0; x_{\pi(A)}) \leq 1$. With these ingredients, we can formulate a double loop IPF for maximum conditional likelihood for joint parent chain graphs. The outer loop is the iteration over $\alpha$, which is also present in IPF for unconditional learning and for conditional learning with clamped parents. The inner loop, which is only present in IPF for maximum conditional likelihood in joint parent chain graphs, is needed for the optimization of $\lambda$, to make sure that the potentials are properly normalized.

**Example.** In a Bayesian network, $\widetilde{\Psi}_{A\backslash\alpha}(x_A) = 1$. Therefore we can simplify

$$\psi_\alpha^\lambda(x_\alpha) = \frac{\widetilde{P}^D(x_\alpha)}{\widetilde{g}_\alpha^c(x_\alpha) + \lambda(x_{\pi(A)})}. \quad (37)$$

## 5 RESULTS

In this section, we apply the IPF procedure to two (toy) problems. The first one is the maximum likelihood estimation in a two-layered sigmoid belief network. In the second one, we fit a Bayesian network to a table of conditional probabilities, obtained from medical literature. In both problems, IPF is compared with state-of-the-art methods.

### 5.1 SIGMOID BELIEF NETWORKS

We considered two-layered sigmoid belief networks with $n = 5$ variables $x_i$ in the top-layer and 5 variables $y_j$ in the bottom layer.

Following [19], we have constructed 100 times an artificial data sets by generating $p = 2n$ random patterns $(x^\mu, y^\mu)$ from a uniform distribution. This choice for $p$ is motivated by the fact that for very small dataset, $p \ll 2n$, or very large ones, $p \sim 2^n$, correlations in the generated data are more or less absent, which makes these data sets less interesting for learning.

For all data sets, maximum likelihood estimation with IPF for sigmoid belief networks has been compared with Polak-Ribiere's conjugate gradient (CG) [18] and steepest descent (SD) applied directly to (minus) the log-likelihood parameterized by the weights and biases. In both CG and SD the gradients of the log-likelihood are computed as a subroutine. These are

$$\frac{\partial L}{\partial w_{ij}} = \langle y_j x_i \rangle_{\text{data}} - \langle \langle y_j \rangle_x x_i \rangle_{\text{data}} \quad (38)$$

$$\frac{\partial L}{\partial h_i} = \langle y_j \rangle_{\text{data}} - \langle \langle y_j \rangle_x \rangle_{\text{data}} \quad (39)$$

where $\langle \ldots \rangle_{\text{data}}$ is the average over the empirical data distribution and $\langle . \rangle_x$ the conditional average according to the current model $P(y|x, w, h)$. The gradient is used to determine the direction of a one dimensional line-search. In all three methods, the probabilities in the top layer were kept at $P(x_i = 1) = 0.5$. (If the top layer is to be learned as well, this will make the experiment more favorable for IPF, since it will immediately find the maximum likelihood solution by reading off frequencies from the data).

Results are plotted in Figure 1. The left panel shows the log-likelihood as a function of the cycle number of a typical run. In IPF, each pairwise potential is visited exactly one time in a cycle and is immediately updated. In CG and SD, one cycle consists of first computing all the gradients, and then doing a full line minimization in the (conjugate) gradient direction. The computational costs for computing all the partial derivatives should be about the same as the costs of one cycle in IPF. Experimentally, we found that the computational costs of one cycle CG and SD is about three times the costs in IPF. This may be due to a costly line minimization subroutine. In the left plot, IPF is initially superior. With increasing number of cycles, IPF remains the best one, but the three methods converge to a similar score. The right panel shows the mean and standard deviation of the difference in log-likelihood of IPF and CG. Initially IPF is significantly better than CG. Asymptotically, there seems to be no significant difference. Theoretically, one would expect CG to be superior. Results obtained with SD (not plotted) are significantly inferior compared to IPF and CG.

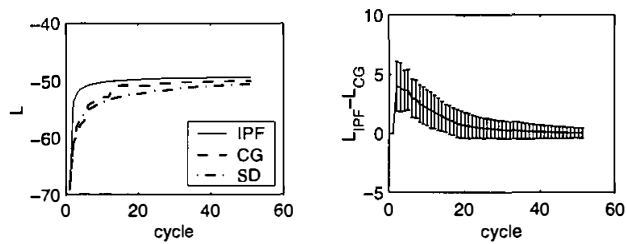

Figure 1: Learning curves for likelihood maximization in sigmoid belief networks. Left: a typical learning curve. Log-likelihood is plotted against cycle number for IPF, CG, and SD. Right: Mean and standard deviation of the difference in log-likelihood between IPF and CG, obtained from 100 simulations with independently generated data sets.

### 5.2 BAYESIAN NETWORKS

We illustrate IPF for conditional maximum likelihood in Bayesian networks by an example involving the diagnosis of coronary heart disease, taken from [20]. It is well known that the prior probability of a patient having coronary heart disease depends on sex and age of the patient. Older men have a much higher probability of having the disease than young women. A typical symptom of this disease is *Angina Pectoris* (chest pain). We have constructed a model $P$ on the basis of information given in [20], which is a biannual publication by the Dutch national health insurances, distributed among all Dutch physicians.

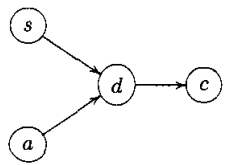

Figure 2: Bayesian network to model coronary heart disease.

In this example, we have four variables: *age* ($a$), *sex* ($s$), *heart-disease* ($d$), and *chest-pain* ($c$). In the example, $a$ has four states (30-39, 40-49, 50-59, 60-69), $s$ has two states (*male, female*), $d$ has two states (*true, false*), and $c$ has four states, (asymptomatic, non-AP pain, atypical AP-pain, typical AP-pain). Assuming that $s$ and $a$ are independent, and that $c$ is independent of $s$ and $a$ given $d$, we have the graphical structure of Figure 2.

In [20] conditional probabilities $Q(d|a,s,c)$ based on data are tabulated (see Table 1). Note that the 'directions' of $Q$ do not match with the 'directions' in the model in Figure 2. With $Q$ – which can be interpreted as the conditional frequencies in a conditional

Table 1: Conditional probabilities of heart disease given sex ($s$), age ($a$) and type of chest-pain (asymptomatic, non-AP pain, atypical AP-pain, typical AP-pain). This table is taken from the literature and serves as conditional data for the probability model in Figure 1.

| $s$ | $a$ | asympt | non-AP | atyp-AP | typ-AP |
|---|---|---|---|---|---|
| m | 30-39 | 0.019 | 0.052 | 0.218 | 0.677 |
| m | 40-49 | 0.055 | 0.141 | 0.461 | 0.873 |
| m | 50-59 | 0.097 | 0.215 | 0.589 | 0.92 |
| m | 60-69 | 0.123 | 0.281 | 0.671 | 0.943 |
| f | 30-39 | 0.003 | 0.008 | 0.042 | 0.258 |
| f | 40-49 | 0.01 | 0.028 | 0.133 | 0.552 |
| f | 50-59 | 0.032 | 0.084 | 0.324 | 0.794 |
| f | 60-69 | 0.075 | 0.186 | 0.544 | 0.906 |

database– we learned the parameters of the Bayesian network using IPF for joint parent chain graphs. We compare the results with Jebara's 'reversed Jensen' algorithm [15], which is an advanced algorithm designed for conditional likelihood maximization for models in the exponential family.

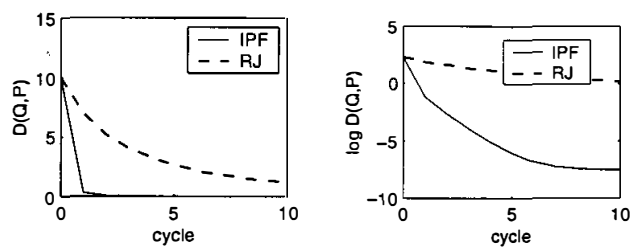

Figure 3: Learning curves for conditional likelihood maximization using IPF and 'Reversed Jensen'

Both algorithms try to minimize the divergence

$$D(Q,P) = \sum_{asdc} Q(d|asc) \log \frac{Q(d|asc)}{P(d|asc)}. \qquad (40)$$

The learning curves are shown in Figure 3. For both algorithms, $D$ and $\log D$ are plotted as a function of cycles respectively in the left and right panel. In both algorithms, all the parameters are updated exactly once during a cycle. In our experiment, we found that computation time per cycle was about the same for both algorithms. Clearly, IPF shows favorable results.

## 6 DISCUSSION

In this paper, we presented IPF-like algorithms for unconditional and conditional likelihood maximization. All these algorithms are based on the principle of maximization via auxiliary functions. These functions have been constructed by employing Jensen's bound, and the linear upper bound of the log. Similar constructions for auxiliary functions for (conditional) log-likelihood with exactly the same bounds have been presented before [16, 21]. However, in e.g. [21], where the construction has been applied for conditional Gaussian mixture modeling, the method leads to rather cumbersome expressions. In [15] better results are reported using the reverse Jensen procedure, which is based on a quadratic bound. In chain factor graphs with discrete variables, the situation is completely different. Thanks to the definition of chain factor graphs as products of local potentials, the auxiliary functions have a conceptually very simple and elegant form, and they are easily maximized in closed form. The solution generalizes upon standard IPF applied to undirected and chain graphs [6, 7]. In addition, simulations show competitive performance compared to other state of the art methods.

## Acknowledgments

This project is funded by the Dutch Technology Foundation STW. We thank Jan Neijt for pointing us to the medical example, the anonymous referees for their useful comments, and David Barber for help.